\documentclass[10pt,conference]{IEEEtran}
\IEEEoverridecommandlockouts
\usepackage{cite}
\usepackage{amsmath,amssymb,amsfonts}
\usepackage{algorithmic}
\usepackage{graphicx}
\usepackage{textcomp}
\usepackage{xcolor}
\usepackage[hidelinks]{hyperref}
\usepackage{listings}
\usepackage[linesnumbered,ruled]{algorithm2e}
\def\BibTeX{{\rm B\kern-.05em{\sc i\kern-.025em b}\kern-.08em
    T\kern-.1667em\lower.7ex\hbox{E}\kern-.125emX}}
\begin{document}

\title{\textit{Fairpriori}: Improving Biased Subgroup Discovery for Deep Neural Network Fairness}

\author{\IEEEauthorblockN{Anonymous authors}

}

\author{
\IEEEauthorblockN{Kacy Zhou\IEEEauthorrefmark{1}, Jiawen Wen\IEEEauthorrefmark{1}, Nan Yang\IEEEauthorrefmark{1}, Dong Yuan\IEEEauthorrefmark{1}, Qinghua Lu\IEEEauthorrefmark{2}, Huaming Chen\IEEEauthorrefmark{1}}
\IEEEauthorblockA{\IEEEauthorrefmark{1}The University of Sydney, Australia}
\IEEEauthorblockA{\IEEEauthorrefmark{2}CSIRO, Australia}
}

\maketitle

\begin{abstract}
While deep learning has become a core functional module of most software systems, concerns regarding the fairness of ML predictions have emerged as a significant issue that affects prediction results due to discrimination. Intersectional bias, which disproportionately affects members of subgroups, is a prime example of this. For instance, a machine learning model might exhibit bias against darker-skinned women, while not showing bias against individuals with darker skin or women. This problem calls for effective fairness testing before the deployment of such deep learning models in real-world scenarios. However, research into detecting such bias is currently limited compared to research on individual and group fairness. Existing tools to investigate intersectional bias lack important features such as \textit{support for multiple fairness metrics}, \textit{fast and efficient computation}, and \textit{user-friendly interpretation}. This paper introduces Fairpriori, a novel biased subgroup discovery method, which aims to address these limitations. Fairpriori incorporates the frequent itemset generation algorithm to facilitate effective and efficient investigation of intersectional bias by producing fast fairness metric calculations on subgroups of a dataset. Through comparison with the state-of-the-art methods (e.g., Themis, FairFictPlay, and TestSGD) under similar conditions, Fairpriori demonstrates superior effectiveness and efficiency when identifying intersectional bias. Specifically, Fairpriori is easier to use and interpret,  supports a wider range of use cases by accommodating multiple fairness metrics, and exhibits higher efficiency in computing fairness metrics. These findings showcase Fairpriori's potential for effectively uncovering subgroups affected by intersectional bias, supported by its open-source tooling at \url{https://anonymous.4open.science/r/Fairpriori-0320}.
\end{abstract}

\begin{IEEEkeywords}
machine learning, fairness, intersectional bias, subgroup discovery
\end{IEEEkeywords}

\section{Introduction}
The prevailing trend of integrating deep neural networks (DNNs) across various industry sectors has demonstrated the immense potential of DNNs, enabling analysis and prediction capabilities in the form of software applications~\cite{zhou2018human,rech2004artificial}. Successful applications include in the fields of finance~\cite{martinez2021secret}, recruitment~\cite{kodiyan2019overview}, justice~\cite{angwin2016machine} and so on. Yet, it is widely acknowledged that machine learning (ML) and deep learning (DL) algorithms are susceptible to learning from historical data, potentially leading to bias and unfair decision-making\cite{chakraborty2021bias,chen2023fairness,chen2022maat}.

Unfair decision-making rooted in the algorithms used generally result because of various attributes in historical data, which will affect functional software systems, such as recommender systems\cite{gomez2021winner}, providing biased solutions to different stakeholders. This effect highlights the significance of mitigating software unfairness within the software systems, from which the terms `unfairness issues' as `fairness bugs' in software engineering are coined~\cite{chen2023fairness}.

In data-driven DL-based systems, fairness aims to ensure that a model's output is independent of individual samples or associated attributes that may introduce unfairness or bias~\cite{oneto2020fairness}. This research area has garnered much attention in recent years, driven by various regulatory and stakeholder requirements. Intersectional bias, which refers to subgroups where members share more than one sensitive attribute, is presented only recently in fairness testing, e.g., TestSGD~\cite{zhang2022testsgd}. It represents a relatively under-explored element of deep learning models, focusing on prejudice against individuals who have multiple protected attributes. For instance, Buolamwini and Gebru demonstrated that commercially available gender classification software systems displayed lower accuracy rates for darker-skinned women when compared to men with lighter skin~\cite{buolamwini2018gender}. 

There is a scarcity of research into intersectional bias in machine learning~\cite{chen2022fairness}, as the vast majority of literature focuses on individual and group fairness rather than subgroup fairness. This is important as simply removing protected sensitive attributes from training a machine learning model with may still result in proxy discrimination amongst other attributes~\cite{corbett2018measure}. This \textit{subgroup discrimination can be subtle, which is difficult to identify}. This illuminates the critical need to effectively and efficiently identify and mitigate subgroup discrimination against sensitive attributes. Furthermore, existing studies \textit{lack support for applying multiple fairness metrics} tailored to different application contexts, which is crucial in deep learning models as inappropriate metrics being applied in a scenario would lead to incorrect conclusions~\cite{cabitza2019wants,hossin2015review}. 
\begin{figure*}[htbp]
    \centering
    \includegraphics[width=.8\textwidth]{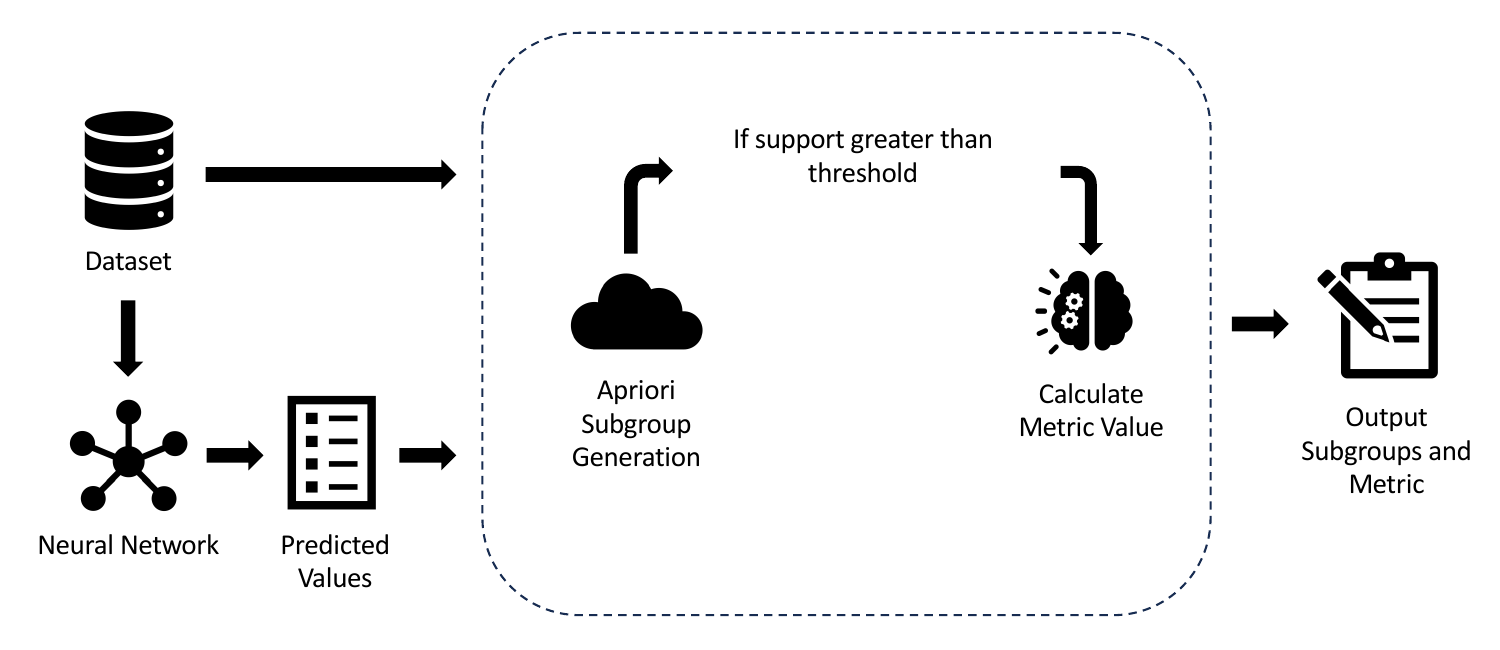}
    \caption{An Overview of Fairpriori}
    \label{fig:fairpriori}
\end{figure*}
Addressing the research gaps in identifying biased subgroups within deep learning systems, this paper presents \textit{Fairpriori} as a novel solution to these issues. \textit{Fairpriori} integrates the calculation of fairness metrics into the process of generating frequent itemsets, which are sets of items that commonly occur together within a database. This integration aims to enhance the efficiency of subgroup generation, enabling the support of a broader range of fairness metrics for measuring various types of fairness. Furthermore, \textit{Fairpriori} incorporates the inclusion of multiple fairness metrics through distinct computations in the numerator and denominator components. This approach allows for the effective and efficient calculation of a final value for each instance subgroup, ensuring precise fairness assessment. We extensively evaluate \textit{Fairpriori} against other state-of-the-art biased subgroup discovery methods, such as Themis~\cite{galhotra2017fairness}, FairFictPlay~\cite{kearns2018preventing}, and TestSGD~\cite{zhang2022testsgd}. The experiment results demonstrate that \textit{Fairpriori} is more efficient and effective, reducing the training time from several minutes to several seconds. It also presents superiori integration due to \textit{Fairpriori}'s more interpretable outputs. 

In summary, \textit{Fairpriori} offers significant advancements to improve biased subgroup discovery in deep learning systems. The contributions of \textit{Fairpriori} are as follows:
\begin{itemize}
    \item We present \textit{Fairpriori} as a novel method designed to automatically identify biased subgroups. It features a variety of parameters with default settings to accommodate diverse application tasks across various data formats.
    \item We innovatively incorporate multiple fairness metrics in \textit{Fairpriori} to support different fairness definitions for different application tasks.
    \item We demonstrate that \textit{Fairpriori} can effectively and efficiently generate samples for a wide range of parameters to mitigate subgroup bias.
    \item We release the open source code of \textit{Fairpriori}, which implements easy-to-use function calls with data inputs and outputs that can be quickly interpretable.
\end{itemize}

The rest of this paper is organised as follows. Background knowledge on subgroup fairness and intersectional bias will be discussed in \autoref{sec:background1}. Next,  the preliminaries of fairness metrics and existing methods for discovering biased subgroups are explained in \autoref{sec:background2}. \textit{Fairpriori} will be introduced in \autoref{sec:fair}, and it will be systematically analysed in \autoref{sec:rqs}. We discuss the threat to validity with conclusion in \autoref{sec:tot} and \autoref{sec:conclusion}, respectively.

\section{Subgroup Fairness and Intersectional Bias}\label{sec:background1}
Subgroup fairness is closely related to the concepts of individual fairness and group fairness. In fairness metrics, individual fairness means that similar individuals should have similar predictions, while group fairness indicates sets of individuals having the same value for one attribute should be treated equally~\cite{binns2020apparent}. For example, when determining college admissions, one could consider individual fairness to be that two candidates with \textit{similar grades}, \textit{extra-curriculars}, and \textit{admission essays} should have the same decision. Meanwhile, group fairness would imply similar decision rates for different genders, such as males and females. In essence, this refers to the granularity of fairness being measured.

However, \textit{subgroup} fairness focuses on groups that consist of the intersection of different groups, embodying a finer granularity than \textit{groups} yet involving more than a single individual. For example, a possible subgroup could be characterised by an intersection of race and gender (such as Asian females). Intersectional bias is therefore concerned with bias affecting those belonging to more than one protected group~\cite{wang2022towards}. 

As mentioned, one example of subgroup fairness is a study by Buolamwini and Gebru \cite{buolamwini2018gender} which found that three different commercial gender classification systems misclassified darker-skinned females with error rates of up to 34.7\%, while for lighter-skinned males the maximum error rate was 0.8\%. Additionally, another study by Guo and Caliskan \cite{guo2021detecting} on bias in word embedding models used for natural language processing, including technologies like ChatGPT, showed that there was subgroup bias against Mexican females. These are troubling results that show an urgent need for enhanced transparency and accountability in machine learning to ensure fairer results for all, especially for those individuals who may experience bias from multiple sources. 

This is important because intersectional bias tends to be overlooked, often due to the presence of \textit{multiple protected attributes} at the same time~\cite{gohar2023survey}, and the resulting bias comes from systems of oppression that interact with each other \cite{wang2022towards}. As noted in a fairness testing survey by Chen et. al. \cite{chen2022fairness}, while intersectional bias is an important field of research, it remains relatively unexplored in literature, citing several studies, such as \cite{cabrera2019fairvis}\cite{tao2022ruler}\cite{zhang2022testsgd} as examples, in comparison to the extensive other studies on individual or group fairness. This disparity underscores a significant gap and presents an opportunity for in-depth research in the area of intersectional bias and biased subgroup discovery.

\section{Preliminaries}\label{sec:background2}
\subsection{Fairness Metrics}
A challenge that may not be immediately obvious is how fairness can be measured in the outcomes of machine learning models across different subgroups. One could gravitate to using accuracy, which is the proportion of correctly predicted outcomes. For example, if a model was 90\% accurate on white males and 50\% accurate on black females, that could be a cause of concern. However, the use of accuracy as a fairness metric would not be appropriate for all circumstances, and it is sometimes not possible for multiple metrics to be satisfied at once according to the impossibility theorem. To address this complexity, \textit{multiple fairness metrics} have been developed to solve this problem, with several key detailed in \autoref{tab:fairnessmetrics}.
\begin{table}[tbp]
\caption{Summary of important fairness metrics\\from \cite{ruf2021towards}, \cite{mehrabi2021survey}, and \cite{verma2018fairness}.}
\begin{center}
\begin{tabular}{|p{1.5cm}|p{6cm}|}
\hline
\textbf{Metric} & \textbf{Definition} \\ \hline
Demographic Parity & All subjects in all groups have an equal probability of being predicted as positive. \\ \hline
Conditional \newline Statistical \newline Parity & All subjects in all groups have an equal probability of being predicted as positive when legitimate factors are controlled for. \\ \hline
Conditional Use Accuracy \newline Equality & The probability that a subject classified as positive is actually positive and the probability that a subject classified as negative is actually negative should be equal to each other for all groups. \\ \hline
Predictive \newline Parity & The probability that a subject classified as positive actually belongs to the positive class (PPV) should be equal for all groups. \\ \hline
Equalised Odds & All groups should have equal true positive rate, and also have equal false positive rate. \\ \hline
Equalised \newline Opportunities & The probability that a subject classified as negative is actually positive (false negative rate) should be equal for all groups. \\ \hline
Predictive Equality & The probability that a subject classified as positive is actually negative (false positive rate) should be equal for all groups. \\ \hline
\end{tabular}
\label{tab:fairnessmetrics}
\end{center}
\end{table}

A notable instance of selecting an inappropriate fairness metric choice occurred with the Correctional Offender Management Profiling for Alternative Sanctions (COMPAS) tool created by Northpointe (now Equivant), which was used to assess the recidivism risk of criminals in some USA states~\cite{angwin2016machine}. In 2016, a nonprofit newsroom, ProPublica, reported that the tool was unfairly biased against black individuals~\cite{angwin2016machine}. Their main findings were that black defendants were twice as likely to be inaccurately assessed as higher risk but not reoffend (\textit{higher false positive rate}), whereas white defendants were more likely to be mislabelled as lower risk (\textit{higher false negative rate}) \cite{angwin2016how}. 

This led to a rebuttal from Northpointe in \cite{dieterich2016compas}, of which its relevant arguments were that they incorrectly used the false positive rate (proportion incorrectly labelled as high risk over all those who did not reoffend) instead of the false discovery rate (proportion incorrectly labelled as high risk over all those labelled as high risk), and their tool satisfied predictive parity instead of equalised odds. They pointed out that, while the base rates of recidivism were different between blacks and whites (0.51 vs. 0.39 for general recidivism), the calculation of false positive rate would be affected, therefore making it an unsuitable metric since any algorithm would classify more blacks compared to whites as higher risk due to this. They suggested that, the absence of bias should be evaluated through the lens of false discovery rate,  under which their tool did not exhibit bias.

It highlights the controversy that ProPublica and Northpointe are evaluating fairness through different measurements of fairness. Due to the impossibility theorem, it is not possible to be simultaneously fair in different definitions at once. However, in this situation, Northpointe's definition of fair was further validated by the third-party think tank Community Resources for Justice~\cite{flores2016false}, which supported Northpointe's findings in \cite{dieterich2016compas}. They also critiqued ProPublica's methodology and analysis due to their failure to use correct methods and standards for bias, with one main point being that COMPAS was created to estimate the probability prediction instead of an absolute determination, which ProPublica did not take into consideration. Because COMPAS was being used as an actuarial risk system, it could only inform on the probabilities of reoffending and cannot be expected to determine absolutely whether an individual will or will not reoffend \cite{flores2016false}.

It is recognised that choosing a fairness metric is a challenging task. Some tools, such as Ruf and Detyniecki's Fairness Compass~\cite{ruf2021towards}, or Aequitas' Fairness Tree from Saleiro et. al.~\cite{saleiro2018aequitas}, have been developed to facilitate this process. These resources offer structured and formalised methods for selecting suitable fairness metrics, ensuring that the choice is methodologically sound and tailored to the specific context.


\subsection{Biased Subgroup Discovery}

Biased subgroup discovery aims to identify subgroups that might be adversely impacted by bias in a machine learning model. For this purpose, we present three particular methods for biased subgroup discovery, namely Themis, FairFictPlay, and TestSGD. These methods are detailed below.

\subsubsection{Themis}
Introduced by Galhotra et al. in 2017, Themis marks one of the earliest tools for identifying subgroups impacted by intersectional bias~\cite{galhotra2017fairness}. It discovers affected subgroups by generating combinations of attributes that discriminated beyond a specific threshold. To achieve this, Themis defines and calculates a group discrimination score as the greatest difference in demographic parity over a subset of attributes and is based upon the Calders-Verwer score which measures the difference between the smallest and largest fractions of outcomes \cite{calders2010three}. To manage its extensive search space, Themis implements several efficiency optimisations, such as caching previous values to be reused if necessary, pruning to reduce the search space, and sampling the search space to reduce the time cost.

One notable aspect of Themis is that an oracle is not required for its operation (actual values as opposed to predicted values). This characteristic is particularly relevant in scenarios where the fairness metric, such as demographic parity, does not necessitate actual values. However, this also means that for fairness metrics requiring real outcomes, Themis might not be the best fit, echoing the considerations raised in discussions about the COMPAS tool's evaluation.

\subsubsection{FairFictPlay}
Developed in 2018, FairFictPlay focuses on auditing subgroup fairness when given access to an oracle that can provide ground truth data~\cite{kearns2018preventing}. The measures of fairness that FairFictPlay supports are equalised opportunities where equality of false negative rates between subgroups is required, and predictive equality where equality of false positive rates between subgroups is required. This tool identifies the members of biased subgroups and calculates a disparity (fairness) score, reflecting the difference in the fairness metrics between the subgroup and the overall population.

FairFictPlay operates by orchestrating a competition between a learner and an auditor within a zero-sum game framework, a process that asymptotically converges to identify unfairness, yet is more straightforward and efficient in computation. The learner's objective is to optimize a function that balances accuracy and fairness for subgroups spotlighted by the auditor, who aims to uncover the subgroups most significantly affecting fairness.

While there is no indication that Kearns et al. were aware of Themis and so they cannot be compared, their work proves effective in detecting impacted subgroups and their disparity scores across four tested fairness datasets. 

\subsubsection{TestSGD}\label{sec:testsgd}
In 2022, TestSGD is one most recent testing approach that aims to identify and measure subgroup discrimination~\cite{zhang2022testsgd}, guiding the new sample generation to mitigate the bias effects during model retraining without sacrificing accuracy.

This method classifies subgroups through linear equality rules for continuous attributes, such as `$\text{age} \geq 30$' or specific values for categorical attributes, like `$\text{gender} = \text{F}$'. A combination of these rules specifies a subgroup, for instance, `$\{\text{age} \geq 30 \wedge \text{gender} = \text{F}\}$'. It accommodates multiclass classification by varying each rule to predict different outcomes, enabling the identification of subgroups like $\{\text{age} \geq 30 \wedge \text{gender} = \text{F}\} \rightarrow \text{positive}$, which describes the subgroup of females who are 30 or older and are predicted to be positive. Subgroup discovery relies on the Apriori algorithm for frequent itemset generation~\cite{apyori}, which identifies common combinations of attributes, thereby defining subgroups. The implementation of Apyori helps pinpoint the largest subgroups early on, streamlining the search process. The fairness of a subgroup is evaluated based on the absolute difference in outcomes between members within and outside the subgroup, focusing on demographic parity as the fairness criterion. This assessment involves sampling to manage computational demands, similarly to Themis, and reports the associated error margin.

To address identified biases, TestSGD applies random adjustments to non-protected attributes of selected samples that meet the subgroup criteria, thereby generating data points for retraining models towards improved fairness.

Zhang et al.'s testing reveals TestSGD's efficacy in uncovering intersectionally biased subgroups within machine learning models, showcasing reasonable efficiency and the ability to lessen such biases through retraining with synthetically generated data. Further research by Guo et al. in 2023 demonstrates TestSGD's superiority in efficiency over Themis by at least a twofold margin~\cite{guo2023fairrec}. TestSGD outperforms both Themis, by adopting a more proactive approach in pruning non-viable subgroups without the need for sampling from an external file, and FairFictPlay, by offering higher interpretability and specificity in identifying biased subgroups through rule-based definitions for subgroups instead of grouping all biased subgroups.

\section{Fairpriori}\label{sec:fair}
While the aforementioned methods have presented tentative solutions for biased subgroup discovery, improving their effectiveness and efficiency remains a daunting task, which poses significant challenges. In this section, we deliberately design \textit{Fairpriori} to enhance both aspects, the code of which is accessible through the anonymous link at \url{https://anonymous.4open.science/r/Fairpriori-0320}. Python 3 was chosen for its implementation, capitalizing on its rich ecosystem of machine learning and data analytics libraries like scikit-learn, Keras, and TensorFlow. This choice facilitates its integration into various workflows as a package through a simple function call. To further advance efficiency, Fairpriori employs a frequent itemset generation algorithm for identifying subgroups, the Apriori algorithm. Innovatively, Fairpriori directly integrates the steps of subgroup generation with the calculation of fairness metrics, streamlining the process by eliminating the need for separate, extensive computations for each subgroup. This method not only boosts efficiency but also simplifies the inclusion of diverse fairness metrics. The implementation details and pseudocode of Fairpriori are provided for a comprehensive overview of its functionality.

\begin{algorithm}[tbp]
    $L_1 = \text{Frequent itemsets of size 1}$\;
    $k = $ 1\;
    \While{\text{\upshape there are still frequent itemsets of size} $k$}{
        $C_{k+1} = $ Candidate itemsets of size $k+1$ generated from $L_k$\;
        \ForEach{\upshape itemset $i$ in $C_{k+1}$}{
            $C_i = $ Count of this itemset\;
            \If{\upshape $C_i$ makes $i$ frequent}{
                $N_i = $ The numerator value of this itemset\;
                $D_i = $ The denominator value of this itemset\;
                $M_i = N_i / D_i$\;
                Add $\{i, (C_i, M_i)\}$ to $L_{k+1}$\;
            }
        }
        Increment $k$ by $1$\;
    }
    $L_0 = \text{Entry representing the whole dataset}$\;
    $L = \cup L_k$\;
    \ForEach{\upshape itemset $i$ in $L$}{
        Add $M_i - M_{L_0}$ to $i$.
    }
    \Return{\upshape $L$}
    \caption{Fairpriori algorithm.}\label{alg:fairpriori}
\end{algorithm}
\subsection{Multiple Fairness Metrics}
We present the Fairpriori algorithm in \autoref{alg:fairpriori}. In the algorithm, $N_i$ and $D_i$ are designed to accommodate the calculation of multiple fairness metrics, which is achieved through breaking down the fairness metric calculations into their numerator and denominator components. Given that all metrics represent a proportion of the dataset determined uniquely by each metric, individual instances contribute either to the numerator, the denominator, or neither. Consequently, the aggregate contributions for each subgroup's instances can be calculated and then divided to yield the metric's final value.

\begin{table}[htbp]

\centering
\caption{An example dataset for calculating predictive parity, where $1$s for the numerator and denominator denote those that will contribute to the calculation of predictive parity.}\label{tab:res_mfm1}
\begin{tabular}{|l|l|l|l|l|l|l|}
\hline
\textbf{Name} & \textbf{Gender} & \textbf{Race} & \textbf{Actual} & \textbf{Predicted} & \textbf{N} & \textbf{D} \\ \hline
Alex          & Female          & Caucasian     & Positive           & Positive        &       1             &        1              \\ \hline
Ben           & Male            & Asian         &     Negative       & Positive        &        0           &       1               \\ \hline
Cam           & Male            & Asian         & Positive           & Positive        &      1              &      1                \\ \hline
Dan           & Male            & Asian     & Negative           & Negative        &         0           &          0            \\ \hline
\end{tabular}

\end{table}
Let's consider a scenario where we are given a dataset in \autoref{tab:res_mfm1}, and our objective is to determine the predictive parity for the subgroup defined by ${\text{gender} = \text{Male} \wedge \text{race} = \text{Asian}}$. Predictive parity refers to the ratio of true positive outcomes to all positive predictions within a subgroup. Observing the entire group under consideration, the predictive parity is 66\%. This calculation involves identifying true positives (in this case, Alex and Cam are true positives) and all positive predictions (Alex, Ben, and Cam fall into this category).

To specifically calculate the predictive parity for the subgroup of Asian males, we look at the relevant instances in the numerator (true positives) and the denominator (all positive predictions) that pertain to this subgroup. Assuming Ben, Cam, and Dan fall under the Asian male subgroup, and considering that Cam is a true positive while both Ben and Cam are predicted as positive, the predictive parity calculation for this subgroup involves dividing the count of true positives by the count of all positive predictions within the subgroup, resulting in a predictive parity of 50\%.



\autoref{tab:res_met} gives the numerator and denominator calculations for each metric according to the definitions in \autoref{tab:fairnessmetrics}.

\begin{table}[htbp]
\caption{How numerators and denominators were calculated for each fairness metric supported by Fairpriori.}
\label{tab:res_met}
\centering
\begin{tabular}{|p{2cm}|p{1.5cm}|p{1.5cm}|p{2cm}|}
\hline
\textbf{Metric}         & \textbf{Definition} & \textbf{Numerator}                 & \textbf{Denominator}            \\ \hline
Demographic Parity      &           Proportion of positive predictions.          & Instances \newline predicted as \newline positive    & All instances                   \\ \hline
Predictive \newline Parity       &      Proportion of \newline correctly predicted positives.               & Instances that are true \newline positives  & Instances \newline predicted as \newline positive \\ \hline
Predictive Equality     &         Proportion of \newline incorrectly predicted positives.            & Instances that are false \newline positives & Instances \newline predicted as \newline positive \\ \hline
Equalised \newline Opportunities &       Proportion of \newline incorrectly predicted negatives.              & Instances that are false \newline negatives & Instances \newline predicted as \newline negative \\ \hline
\end{tabular}

\end{table}
\subsection{Usage}
To cater to a broad spectrum of applications, the Fairpriori algorithm is configured with inputs that serve specific purposes and come with predefined default settings, as detailed in \autoref{tab:res_mfm}. These settings include a minimum support threshold set at 50\%, meaning that only subgroups constituting a majority of the dataset are considered significant enough to be returned. This ensures focus on substantial segments of the data. Additionally, the maximum length of a subgroup, defined by the number of attribute-value pairs, is set to 2 to provide easily interpretable results. This constraint is intended to keep the results straightforward and easy to understand.

The algorithm also requires specifying the range of possible values for both predicted and actual outcomes, alongside the target label for which calculations will be performed. The default fairness metric selected is demographic parity. This choice is due to its independence from ground truth values, making it straightforward to interpret based on the rate of positive predictions. Furthermore, Fairpriori offers the flexibility to exclude certain attributes from subgroup generation. This feature allows users to prevent the algorithm from considering specific attributes in its analysis, though a simpler method might be to remove these attributes from the data frame before processing.



\begin{table}[tbp]
\centering
\caption{List of Fairpriori inputs, their purpose, and their default value if applicable.}
\begin{tabular}{|p{2.5cm}|p{2.5cm}|p{2cm}|}
\hline
\textbf{Input}            & \textbf{Purpose} & \textbf{Default Value} \\ \hline
Pandas Dataframe of the Dataset                 &      Subgroup generation.                & n/a, mandatory         \\ \hline
Predicted Values          &      Metric calculation.                & n/a, mandatory         \\ \hline
Actual Values             &     Metric calculation.                 & n/a, mandatory         \\ \hline
Value Categories             &     Metric calculation.                 & n/a, mandatory         \\ \hline
Positive Target Label     &           Metric calculation.           & n/a, mandatory         \\ \hline
Minimum Support Threshold &         Subgroup generation.             & 50\%                   \\ \hline
Maximum Length            &        Subgroup generation.              & 2                      \\ \hline
Fairness Metric           &       Metric calculation.               & Demographic Parity     \\ \hline
Ignored Attributes        &        Subgroup generation.              & None                   \\ \hline
\end{tabular}

\label{tab:res_mfm}
\end{table}

A dataframe with four columns is returned from Fairpriori, which contains the subgroups generated defined by a set of `attribute=value' pairs, the support calculated for that subgroup, the metric value calculated for that subgroup, and finally the difference between the metric value for the subgroup and the metric value for the whole dataset. The difference value is positive if the subgroup metric has a higher value than the dataset metric, and negative if it is lower. While not technically a subgroup, the metric value for the whole dataset is returned in the first row in the dataframe, with label `All' and support 1, so that the value is accessible.

\begin{figure}[tbp]
    \centering
    \includegraphics[width=0.45\textwidth]{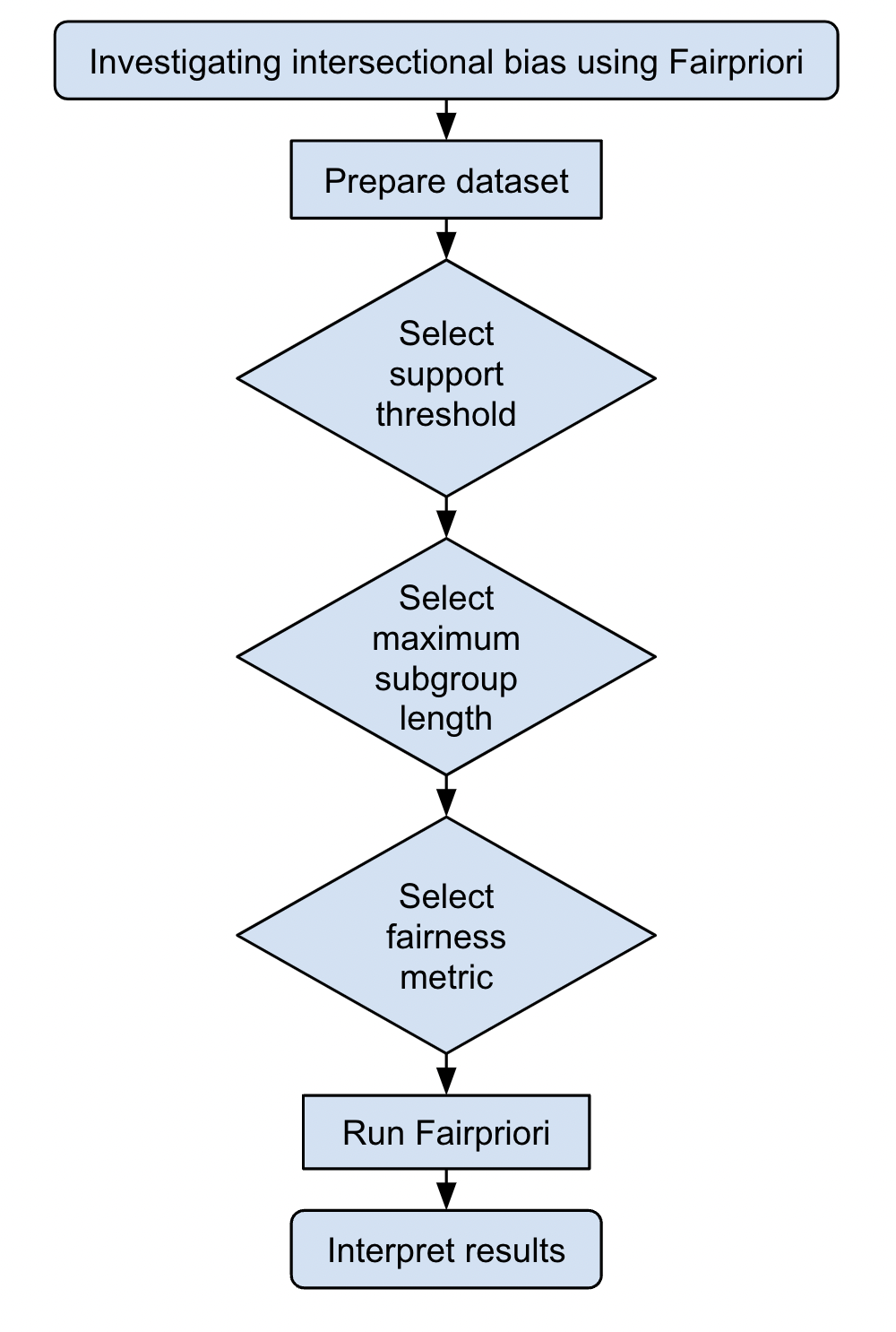}
    \caption{Straightforward flowchart of Fairpriori Application}
    \label{fig:flow}
\end{figure}

The process for employing Fairpriori is illustrated through a straightforward flowchart in \autoref{fig:flow}. Further example usages of Fairpriori are at \url{https://anonymous.4open.science/r/Fairpriori-0320}.


\subsection{Case Study}

A toy machine learning model was developed using the COMPAS dataset, which is now used to examine whether the model's outcomes exhibit bias towards any particular subgroups. With the list of predicted outcomes and actual values from the model, alongside the positive target label indicating a predicted low risk of reoffending, Fairpriori returns the results in \autoref{tab:efu_1} as an output.



\begin{table}[htbp]
\caption{Dataframe results from using Fairpriori with default settings}
\centering
\begin{tabular}{|l|l|l|l|}
\hline
\textbf{itemsets}       & \textbf{support} & \textbf{metric} & \textbf{difference} \\ \hline
(All)                   & 1                & 0.558328        & 0                   \\ \hline
(age\_cat=25 - 45)      & 0.572262         & 0.613817        & 0.055489            \\ \hline
(race=African-American) & 0.51442          & 0.376063        & -0.182265           \\ \hline
(sex=Male)              & 0.809624         & 0.544727        & -0.013601           \\ \hline
\end{tabular}
\label{tab:efu_1}
\end{table}

Because the results haven't produced any subgroups, the support threshold is decreased to 25\% by adding $support=0.25$. \autoref{tab:efu_2} is the result.


\begin{table}[htbp]
\caption{Dataframe results using Fairpriori with default settings except for the minimum support threshold set to 25\%.}
\centering
\begin{tabular}{|p{4cm}|p{1cm}|p{1cm}|p{1.2cm}|}
\hline
\textbf{itemsets}                         & \textbf{support} & \textbf{metric} & \textbf{diff.} \\ \hline
(All)                                     & 1                & 0.558328        & 0              \\ \hline
(age\_cat=25 - 45)                        & 0.572262         & 0.613817        & 0.055489       \\ \hline
(race=African-American)                   & 0.51442          & 0.376063        & -0.182265      \\ \hline
(race=Caucasian)                          & 0.340732         & 0.70661         & 0.148282       \\ \hline
(sex=Male)                                & 0.809624         & 0.544727        & -0.013601      \\ \hline
(age\_cat=25 - 45, race=African-American) & 0.307518         & 0.465753        & -0.092575      \\ \hline
(age\_cat=25 - 45, sex=Male)              & 0.460629         & 0.595146        & 0.036818       \\ \hline
(race=African-American, sex=Male)         & 0.42547          & 0.364052        & -0.194276      \\ \hline
(sex=Male, race=Caucasian)                & 0.262638         & 0.697101        & 0.138773       \\ \hline
\end{tabular}
\label{tab:efu_2}
\end{table}

Concerningly, African American males are much more likely to be predicted as reoffenders with a difference of 19\% to the baseline. However, these results are for the demographic parity fairness metric, which computes the proportion of predicted positives. In the dataset it is more likely for African-Americans to reoffend, and so demographic parity is not an appropriate fairness metric to use in this instance. According to the Fairness Compass \cite{ruf2021towards}, predictive parity is the fairness metric that should be used instead, which measures the probability that a subject classified as positive actually belongs to the positive class. Rerunning the score calculation function in Fairpriori with $metric=``predictive\_parity''$ and the correct settings, \autoref{tab:efu_3} is generated.


\begin{table}[htbp]
\caption{Dataframe results using Fairpriori with default settings except for the minimum support threshold set to 25\% and the demographic parit fairness metric set to predictive parity.}
\label{tab:efu_3}
\centering
\begin{tabular}{|p{4cm}|p{1cm}|p{1cm}|p{1.2cm}|}
\hline
\textbf{itemsets}                         & \textbf{support} & \textbf{metric} & \textbf{difference} \\ \hline
(All)                                     & 1                & 0.767266        & 0                   \\ \hline
(age\_cat=25 - 45)                        & 0.572262         & 0.733856        & -0.03341            \\ \hline
(race=African-American)                   & 0.51442          & 0.695142        & -0.072124           \\ \hline
(race=Caucasian)                          & 0.340732         & 0.796097        & 0.028831            \\ \hline
(sex=Male)                                & 0.809624         & 0.76194         & -0.005327           \\ \hline
(race=African-American, age\_cat=25 - 45) & 0.307518         & 0.66629         & -0.100977           \\ \hline
(sex=Male, age\_cat=25 - 45)              & 0.460629         & 0.730496        & -0.03677            \\ \hline
(sex=Male, race=African-American)         & 0.42547          & 0.675732        & -0.091534           \\ \hline
(sex=Male, race=Caucasian)                & 0.262638         & 0.80885         & 0.041583            \\ \hline
\end{tabular}

\end{table}

The findings indicate a higher probability of African-American men being classified as likely to reoffend, with a 9\% disparity compared to the overall population. Moreover, young African-Americans are similarly more prone to being predicted as reoffenders, with a 10\% difference. Armed with this insight from Fairpriori, the model can be refined to enhance fairness for these specific subgroups. Alternatively, they may assess that the observed discrepancy in fairness across subgroups falls within acceptable limits.


\section{Implementation and Evaluation}\label{sec:rqs}
In this work, we design our experiments based on the benchmark datasets of COMPAS~\cite{angwin2016machine} and Diabetes Hospitals dataset~\cite{misc_diabetes_130-us_hospitals_for_years_1999-2008_296}. The datasets have been widely used for evaluating fairness testing in literature~\cite{dixon2018measuring,galhotra2017fairness,zhang2021automatic,chen2024fairness}. We focus on compare Fairpriori with the state-of-the-art methods as the baselines, including Themis~\cite{galhotra2017fairness}, FairFictPlay~\cite{kearns2018preventing}, and TestSGD~\cite{zhang2022testsgd}. We construct fully connected deep neural network models as the backbone for our evaluation. Specifically, we ask the following research questions:
\begin{itemize}
    \item \textbf{RQ1: Effectiveness of Fairpriori:} is our method effective in identifying biased subgroups for deep learning models?
    \item \textbf{RQ2: Efficiency of Fairpriori:} is our method efficient at biased subgroup discovery for deep learning models?
\end{itemize}

\subsection{RQ1: Is Fairpriori Effective?}

The first research question formulated to determine whether Fairpriori is fit for use is whether Fairpriori is effective. To address this, we conduct an in-depth comparison of Fairpriori's performance and user-friendliness relative to Themis, FairFictPlay, and TestSGD. This evaluation encompasses an analysis of function definitions, including the necessary inputs and the nature of the outputs, to identify the practical strengths and weaknesses of each tool.

\subsubsection{Comparison Against Themis}\label{sec:rq1_themis} 
Different from Fairpriori, for Themis, an XML file is firstly needed describing the dataset's protected attributes and the test that will run, for example, it could be the attributes of sex and age. Following this, they need to prepare a second Python file to sample from the dataset described. An example result is: 
\begin{lstlisting}
Input(s): ('sex', 'age_cat') ---> 59.8%
\end{lstlisting} 
From this, we can tell that there is a large disparity in the outcomes (59.8\%) between subgroups defined by \textit{gender} and \textit{age}.

It highlights a \textit{significant improvement of operation} between Themis and Fairpriori regarding input. Setting up Themis involves substantial effort: it requires crafting a dedicated XML file to detail the dataset's sensitive attributes and designate tests for each potential combination of these attributes. Additionally, Themis necessitates a separate script for sampling based on specific attribute combinations, obliging users to write additional code for querying the dataset. For instance, to sample predictions for Caucasian males, Themis might execute a command like `python3 yourfilename.py male caucasian'. On the other hand, Fairpriori simplifies the process by utilizing a dataframe — a standard tool in data analysis, along with a list of predicted and actual values from that dataframe. Furthermore, Fairpriori enhances user flexibility by allowing the selection of a fairness metric and adjusting the size of the generated subgroups.


The second main improvement is in \textit{the output for biased subgroup discovery}. Themis provides a singular value for each test, derived from a particular combination of the protected attributes. This value represents the disparity in positive prediction rates across all subgroups defined by these attributes, specifically, the gap between the highest and lowest fractions of positive outcomes. In contrast, Fairpriori offers a more granular breakdown, returning the fractions associated with the selected fairness metric for every identified biased subgroup. This includes detailed information on which attribute values are subject to bias, rather than merely indicating the presence of bias within certain attributes. Moreover, while Themis employs sampling, introducing an element of confidence and potential error in its outputs, Fairpriori bypasses sampling for a direct calculation, providing an absolute metric value across the dataset. Consequently, Fairpriori is considered more effective, offering benefits like seamless integration into existing workflows, richer detail in output metrics, and the provision of absolute, rather than estimated, metric values.


\subsubsection{Comparison Against FairFictPlay}\label{sec:rq1_ffp}
To continue with FairFictPlay in the workflow for biased subgroup discovery, we will need to decide on the specific type of fairness concern they wish to investigate. Following we discuss on false negatives due to an interest in disparities among those predicted as low risk to reoffend, with following settings:
\begin{lstlisting}
auditor = Model.Auditor(dataframe, actual_values, "FN")
[violated_group, fairness_violation] = auditor.audit(predicted_values)
\end{lstlisting}
violated\_group returned a list of 1s and 0s, which represents all the instances in an affected subgroup, and fairness\_violation returned 0.085, which represented that there is a disparity of 8.5\% in the outcomes of the instances classified as 1 in the list versus the baseline of the entire population.

One main improvement is in \textit{the output for biased subgroup discovery}. Firstly, FairFictPlay returns a list of 1s and 0s the length of the original data input where the 1s represent those belonging to affected subgroups, as well as a percentage detailing the disparity of the group from the baseline for that metric (either FPR or FNR). In comparison, Fairpriori generates a list of subgroups and their fairness metric calculated as well as its difference from the baseline. This is easier to interpret than the list from FairFictPlay. Fairpriori supports the same metrics as FairFictPlay and also two additional ones, namely \textit{demographic parity} and \textit{predictive parity}, allowing for a wider range of use cases. Both are found to be easy to integrate into existing machine learning pipelines, requiring the original dataset as well as actual and predicted values and any settings (e.g fairness metric to be used).

An interesting benefit of FairFictPlay is its ability to incorporate subgroup fairness directly into the training of machine learning models, where Fairpriori doesn't align as effectively. The computational demands of Fairpriori during model training could significantly extend the process, making it less practical for real-time analysis. However, Fairpriori outshines FairFictPlay in terms of the interpretability of its results and the range of supported fairness metrics, making it more accessible and user-friendly for those seeking to conduct fairness analyses.


\subsubsection{Comparison Against TestSGD}\label{sec:rq1_testsgd}
To compare Fairpriori with TestSGD, we firstly use TestSGD to find applicable subgroups, which are termed as rules here. Subsequently, we specify the support threshold, define the categories for outputs, and set the maximum length for a rule that describes a dataset.
\begin{lstlisting}
itemsets = TestSGD.run_apriori(dataframe, support) 
list_of_rules = TestSGD.create_rules(itemsets, output_categories, maximum_length)
\end{lstlisting}
Following this, we iterate over each identified rule to calculate and ascertain its corresponding fairness score and value.
\begin{lstlisting}
testsgd_results = {}
for rule in list_of_rules:
    rule_results =  TestSGD.group_fairness_score(predicted, dataframe, 
        rule, error_threshold, output_categories)
    print(rule, rule_results[0]], rule_results[1])
\end{lstlisting}
For instance, a specific rule derived from COMPAS dataset might state, \textit{``If sex == Male and race == African-American, then Low (Chance of reoffending)''}. The evaluation of this rule could result in a fairness score of 0.33, with an associated margin of error at 0.04. It suggests that the likelihood of an African-American male being predicted as low risk for reoffending varies by 29\% to 37\% compared to individuals outside this subgroup. It's important to note that the positive nature of the value returned does not immediately clarify the direction of bias; further analysis is required to understand the bias's orientation.



While their usage is very similar to each other, one major difference between Fairpriori and TestSGD was their different result interpretations, both for their rules, and also for the metrics returned. For TestSGD, each rule had a version for each possible prediction value, and the score returned is the absolute difference between the probability that an instance satisfying the rule is predicted as that value and the probability that an instance not satisfying the rule is predicted as that value. For example, for the rule ``If race == Caucasian and medicaid == False then Yes'' returned from the Diabetes dataset, it achieved a score of 0.6\% with an error of $\pm$1\%. This effectively means that there was no bias for this subgroup. However, if there was a greater value, the absence of a sign means that it was not possible to tell whether a subgroup was being favoured by the bias or not. 

For Fairpriori, the metric provided for each subgroup directly reflects the proportion of instances within that subgroup aligning with the chosen fairness metric. A key advantage of this approach is the absence of sampling, which eliminates the possibility of error in the metric calculation. However, a notable limitation is Fairpriori's binary classification framework, recognizing only two prediction outcomes: positive and not positive. This contrasts with TestSGD's capability to handle multiclass classification scenarios.

Furthermore, Fairpriori supports a wider array of fairness metrics beyond demographic parity, offering versatility across various fairness considerations. In contrast, TestSGD has been adapted to leverage its findings for enhancing fairness within neural network models through retraining. Both methodologies exhibit equal effectiveness in their respective domains but differ in their result interpretations and benefits, making each more suitable for specific types of fairness evaluations and adjustments in machine learning models.



\subsubsection{Answer to RQ1}

Fairpriori stands on par with, or even surpasses, other cutting-edge methods in terms of effectiveness, offering straightforward result interpretation and additional features like enhanced analysis of multiple fairness metrics. This adaptability enables broader application possibilities and facilitates further enhancements more readily than existing alternatives. It distinguishes itself from Themis by eliminating the need for extensive setup, offering more precise and actionable outcomes compared to FairFictPlay, and delivering results without the uncertainty of associated errors, unlike TestSGD. Although Fairpriori currently does not support multiclass classification or direct improvements to model fairness, these limitations present opportunities for future research and development, positioning Fairpriori as a strong candidate for advancing fairness in machine learning.

\subsection{RQ2: Is Fairpriori Efficient?}\label{sec:rq2}
To assess Fairpriori's performance in terms of speed and efficiency, a second research question focused on evaluating its computational effectiveness. For this evaluation, two datasets were selected based on their appropriate size and the variety of protected attributes available, providing a solid foundation for comparing the efficiency of different subgroup discovery methods. The use of larger datasets facilitates a clearer comparison of algorithm efficiency, as differences in processing time become more pronounced with more data. The COMPAS dataset was primarily selected for its frequent citation in studies concerning fairness in machine learning~\cite{angwin2016machine}. The Diabetes Hospitals dataset~\cite{misc_diabetes_130-us_hospitals_for_years_1999-2008_296}, with its considerably larger volume of data points (101,762, in contrast to COMPAS's 6,171) and a greater number of attributes (20 versus COMPAS's 5), was chosen to enable a comprehensive efficiency benchmarking of subgroup discovery methods.



In all evaluations, except where noted, the following variables were adjusted to explore their impact on the algorithms' efficiency. These adjustments aimed to understand how variations in the dataset size (number of entries) and complexity (number of attributes) influence efficiency. The minimum support threshold was systematically altered across a broad spectrum (from 10\% to 90\%, in 10\% increments) to assess its effect on computational efficiency. Additionally, the length of subgroups, defined by the number of attribute/value pairs, was restricted to 1, 2, and 3 to balance between interpretability and utility, as longer subgroup definitions tend to be less interpretable and useful.

To establish a benchmark, Fairpriori's efficiency was measured against that of Themis, FairFictPlay, and TestSGD, situating it within the context of existing methods for biased subgroup discovery. Themis was evaluated using demographic parity as the common fairness metric, with the COMPAS dataset serving as the basis due to its inability to process larger datasets efficiently. Conversely, the Diabetes dataset tested the capabilities of FairFictPlay and TestSGD, pushing their analytical boundaries. FairFictPlay's assessment involved metrics of predictive equality and equalized opportunities to ensure a fair comparison with Fairpriori. TestSGD's analysis similarly employed demographic parity as the comparative fairness metric. Comprehensive test procedures and the capability for replication are accessible via the provided codebase at \url{https://anonymous.4open.science/r/Fairpriori-0320}.



\subsubsection{Comparison Against Themis}
Themis displayed significantly lower efficiency compared to Fairpriori, even when analysing a relatively small dataset like COMPAS, which consists of 6,171 entries and 3 protected attributes. In the most computationally demanding scenario, characterised by a 10\% support threshold and a maximum subgroup length of 3, Themis required 330.66 seconds to complete its analysis, whereas Fairpriori completed the same task in 0.068 seconds.

The efficiency gap can be attributed to Themis's reliance on external inputs to sample data for each protected attribute combination. Without specific optimisations, this approach necessitates repeatedly reading and querying the dataframe to generate results, significantly increasing computational overhead. Integrating these processes into a single file could enhance Themis's efficiency. Conversely, Fairpriori's design allows it to process data in a single pass without the need for external data, contributing to its superior efficiency.



\subsubsection{Comparison Against FairFictPlay}

\begin{figure}[htbp]
    \centering
    \includegraphics[width=0.5\textwidth]{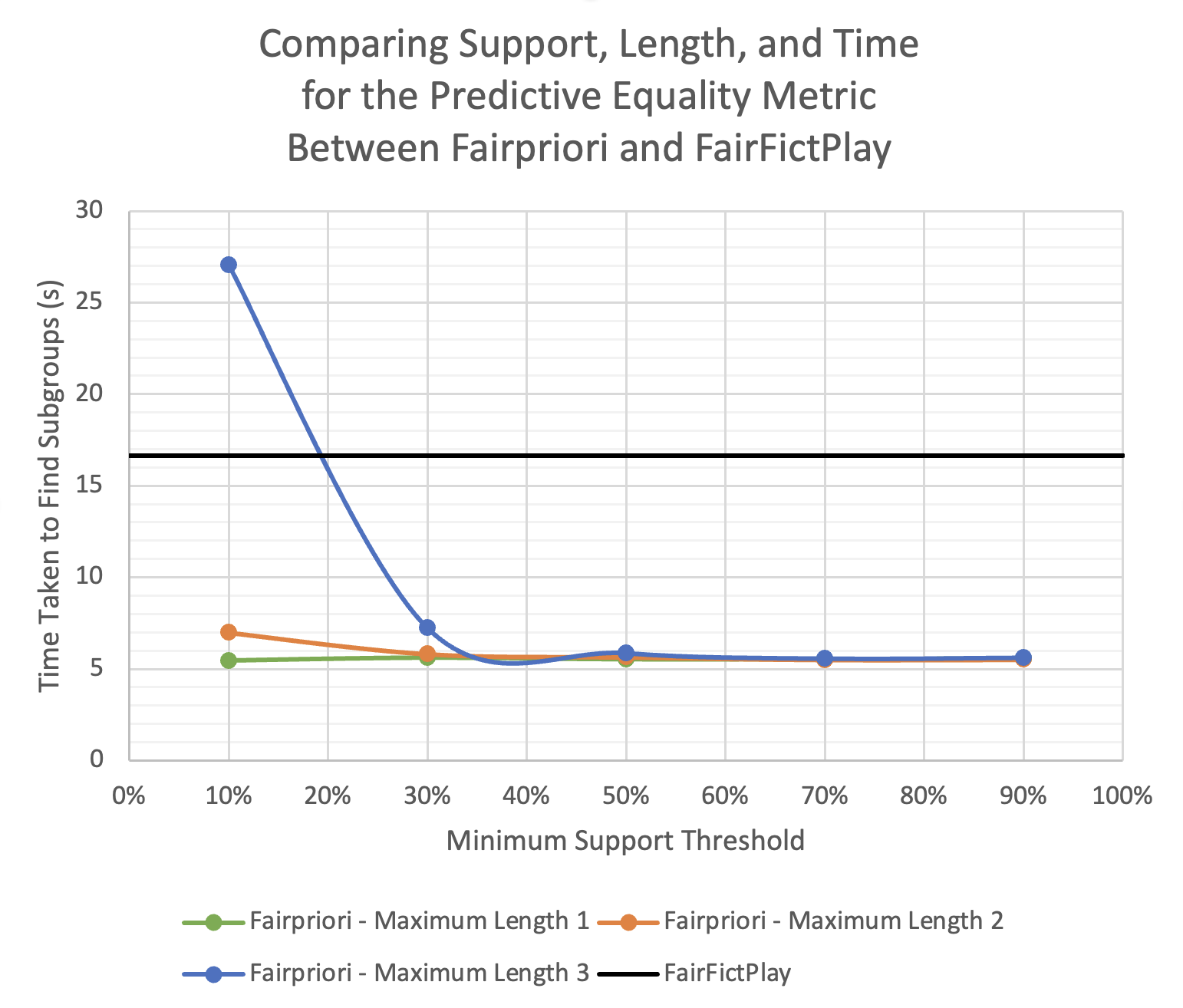}
    \caption{Results comparing the effect of minimum support threshold and maximum subgroup length on time for the predictive equality fairness metric between Fairpriori and FairFictPlay using the Diabetes dataset.}
    \label{fig:fairfictplay_PE}
\end{figure}

The comparison of Fairpriori and FairFictPlay, specifically using the predictive equality metric, as shown in \autoref{fig:fairfictplay_PE}, highlights Fairpriori's significant advantage in efficiency across most tested scenarios, as depicted in the referenced figure. One exception occurs under the most stringent conditions, at 10\% support threshold and a subgroup length of 3, where Fairpriori's processing time is 27.06 seconds, compared to FairFictPlay's 16.66 seconds. For the rest of the settings, Fairpriori consistently outperforms FairFictPlay, often requiring less than half the time to complete the same tasks.

It's anticipated that under more demanding scenarios and with larger datasets like the Diabetes dataset, Fairpriori's performance might not be as robust, primarily due to the exponential increase in the number of subgroups generated by aggressive support and length settings. For instance, a 10\% support with a maximum length of 3 already yields 2,783 subgroups, a number that would surge exponentially with even tighter parameters. However, such aggressive settings would be unlikely to happen in practice. Given that FairFictPlay operates on a polynomial time complexity and Apriori algorithm, which underpins Fairpriori, functions in exponential time, Fairpriori is likely to lag behind FairFictPlay when dealing with larger datasets and more stringent criteria. However, Fairpriori maintains a distinct edge in smaller datasets or when the support and length requirements are less stringent, showcasing its utility in a broad range of practical applications.


\begin{figure}[htbp]
    \centering
    \includegraphics[width=0.5\textwidth]{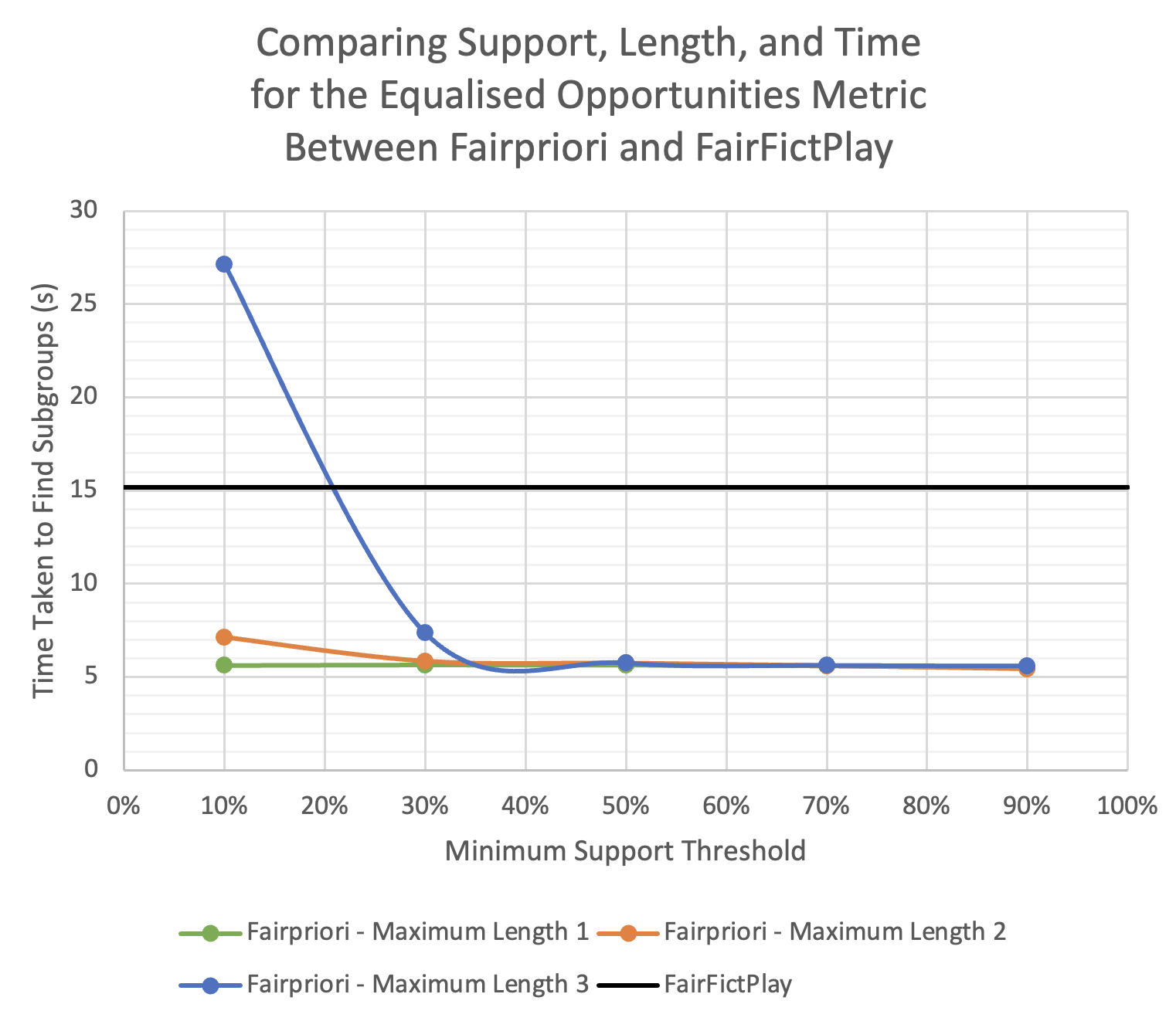}
    \caption{Results comparing the effect of minimum support threshold and maximum subgroup length on time for the equalised opportunities fairness metric between Fairpriori and FairFictPlay using the Diabetes dataset.}
    \label{fig:fairfictplay_EO}
\end{figure}

The analysis using the equalized opportunities fairness metric further corroborates Fairpriori's enhanced efficiency across both metrics, as seen in \autoref{fig:fairfictplay_EO}. Consistently, Fairpriori demonstrates superior efficiency in nearly all tested scenarios.

\subsubsection{Comparison Against TestSGD}

\begin{table}[tbp]
\centering
\caption{Comparing Fairpriori and TestSGD on the Diabetes dataset. Times are rounded 2dp for Adult and Diabetes, and bolded times show the smaller of the two.}
\label{tab:testsgd_efficiency}
\begin{tabular}{|l|l|ll|}
\hline
\textbf{Support} & \textbf{Maximum} & \multicolumn{2}{c|}{\textbf{Diabetes}}                      \\ \cline{3-4} 
\textbf{}        & \textbf{Length}  & \multicolumn{1}{l|}{\textbf{Fairpriori}} & \textbf{TestSGD} \\ \hline
90\%             & 1                & \multicolumn{1}{l|}{5.59s}               & \textbf{5.48s}   \\ \hline
90\%             & 2                & \multicolumn{1}{l|}{5.54s}               & \textbf{5.46s}   \\ \hline
90\%             & 3                & \multicolumn{1}{l|}{5.60s}               & \textbf{5.52s}   \\ \hline
70\%             & 1                & \multicolumn{1}{l|}{\textbf{5.55s}}      & 6.78s            \\ \hline
70\%             & 2                & \multicolumn{1}{l|}{\textbf{5.59s}}      & 8.73s            \\ \hline
70\%             & 3                & \multicolumn{1}{l|}{\textbf{5.60s}}      & 9.14s            \\ \hline
50\%             & 1                & \multicolumn{1}{l|}{\textbf{5.58s}}      & 10.13s           \\ \hline
50\%             & 2                & \multicolumn{1}{l|}{\textbf{5.63s}}      & 22.81s           \\ \hline
50\%             & 3                & \multicolumn{1}{l|}{\textbf{5.79s}}      & 32.91s           \\ \hline
30\%             & 1                & \multicolumn{1}{l|}{\textbf{5.52s}}      & 31.42s           \\ \hline
30\%             & 2                & \multicolumn{1}{l|}{\textbf{5.83s}}      & 79.87s           \\ \hline
30\%             & 3                & \multicolumn{1}{l|}{\textbf{7.50s}}      & 171.66s          \\ \hline
10\%             & 1                & \multicolumn{1}{l|}{\textbf{5.66s}}      & 2,759.37s        \\ \hline
10\%             & 2                & \multicolumn{1}{l|}{\textbf{7.25s}}      & 2,904.08s        \\ \hline
10\%             & 3                & \multicolumn{1}{l|}{\textbf{28.72s}}     & 3,518.86s        \\ \hline
\end{tabular}

\end{table}

\begin{figure}[htbp]
    \centering
    \includegraphics[width=0.5\textwidth]{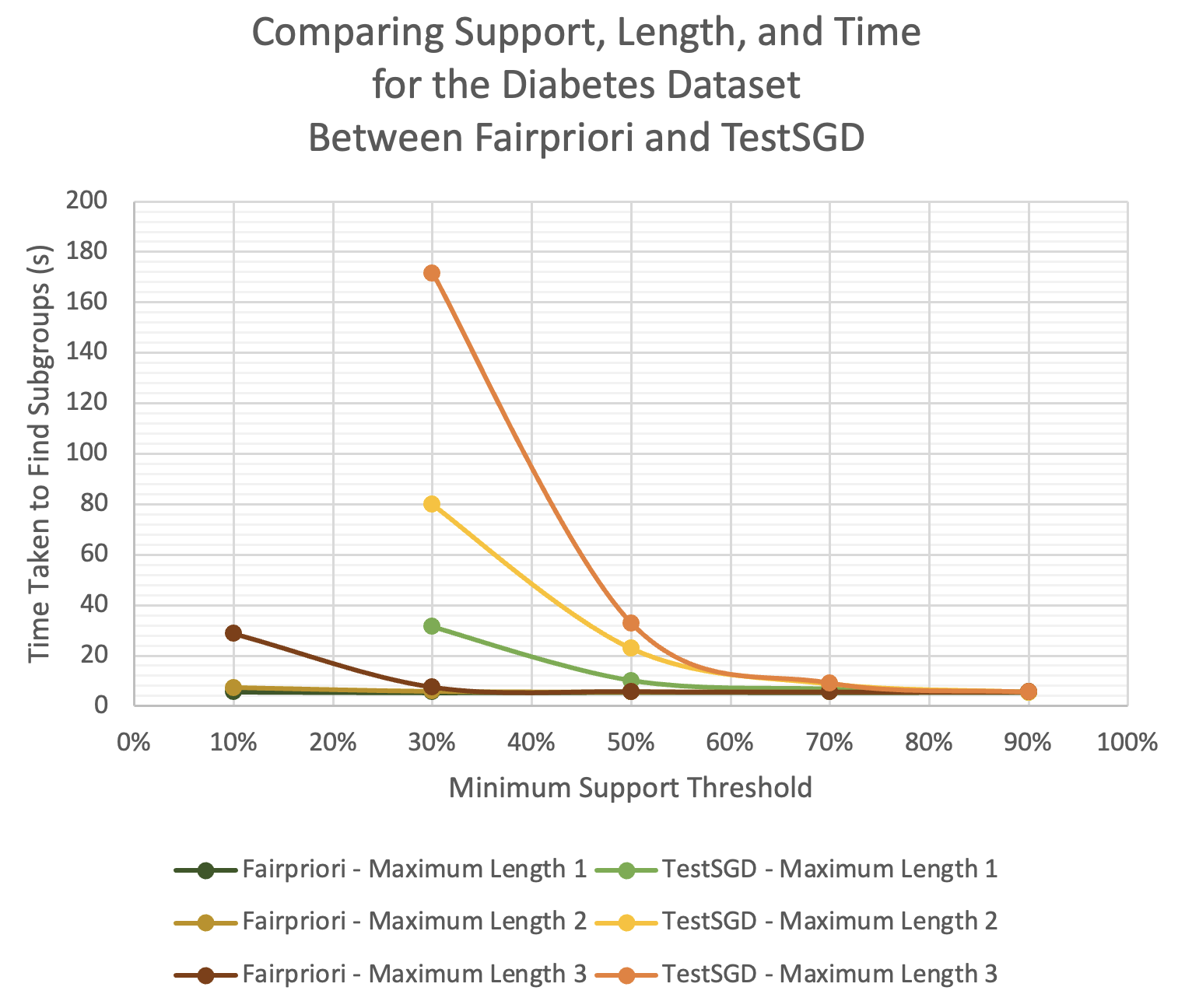}
    \caption{Scatter plot comparing the effect of minimum support threshold and maximum subgroup length on time for the Diabetes dataset between Fairpriori and TestSGD. Results for TestSGD at a support of 10\% are not shown as they took at least 45 minutes and would not be plottable.}
    \label{fig:testsgd_diabetes}
\end{figure}

When investigating the efficiency improvement for the Diabetes dataset in \autoref{tab:testsgd_efficiency} and \autoref{fig:testsgd_diabetes}, there is a clear and large improvement as soon as a 50\% support threshold is reached, as TestSGD starts to head off the plateau. TestSGD at a 10\% support threshold which can not be plotted in \autoref{fig:testsgd_diabetes} takes at least 45 minutes to complete calculations, while Fairpriori takes at most 30s.

These findings highlight Fairpriori's substantial efficiency advantages over TestSGD, primarily due to its method of directly calculating metrics in \autoref{alg:fairpriori}, eliminating the need for sampling. Furthermore, Fairpriori optimises performance by conducting a single set of calculations for each rule, as opposed to TestSGD's approach of evaluating 2 within and outside a given rule or subgroup. 

\subsection{Answer to RQ2}
Fairpriori's efficiency outperforms Themis, FairFictPlay, and TestSGD, all state-of-the-art biased subgroup discovery algorithms, across most scenarios. With Themis, Fairpriori's efficiency advantage is particularly pronounced for the COMPAS dataset, the simplest dataset explored, reducing execution times from minutes to under a second. Against FairFictPlay, Fairpriori's exponential time algorithm generally surpasses FairFictPlay's polynomial time solution, except in cases where the subgroup generation scales to the thousands—an impressive achievement for Fairpriori. When pitted against TestSGD, Fairpriori shows comparable performance in less stringent conditions, but significantly outperforms TestSGD in more complex scenarios, cutting down processing times from over 45 minutes to less than a minute.

These outcomes present Fairpriori as an efficient solution for biased subgroup discovery. This efficiency is largely attributed to the innovative incorporation of fairness metrics directly within Apriori (\autoref{alg:fairpriori}), streamlining the discovery process and significantly enhancing performance.


\section{Threat to Validity}\label{sec:tot}
Despite our efforts for Fairpriori, several potential threats to its validity still exist. Firstly, more user studies could be considered to capture a broader and deeper understanding of how deep learning models are used in practice, and identify areas of Fairpriori that could be improved in response. For example, it may be improved with a wider range of fairness metrics, automated fairness report generation, or a UI interface that could be used by non-technical personnel. Secondly, while Fairpriori shows improvements for the specific datasets and parameters used, further insight with more scenarios, including with more types of ML models, larger datasets, and more demanding parameters, would assist in further characterising its performance in more demanding and varied circumstances.

\section{Conclusion}\label{sec:conclusion}
In this paper, we present Fairpriori, a state-of-the-art method in the field of biased subgroup discovery, which in recent years has been ignored in favour of individual and group fairness research. We consider intersectional bias as a hidden yet critical issue needing an effective and efficient solution. Fairpriori introduces a modified version of a frequent itemset generation algorithm to facilitate its investigation. It effectively incorporates multiple fairness metrics in its analysis, and it is thoroughly evaluated against cutting-edge methods such as Themis, FairFictPlay and TestSGD. We anticipate that our work provides a new frontier for fairness AI research.

\bibliographystyle{ieeetr}
\bibliography{refs.bib}

\begin{thebibliography}{10}

\bibitem{zhou2018human}
J.~Zhou and F.~Chen, {\em Human and machine learning}.
\newblock Springer, 2018.

\bibitem{rech2004artificial}
J.~Rech and K.-D. Althoff, ``Artificial intelligence and software engineering: Status and future trends,'' {\em KI}, vol.~18, no.~3, pp.~5--11, 2004.

\bibitem{martinez2021secret}
E.~Martinez and L.~Kirchner, ``The secret bias hidden in mortgage-approval algorithms,'' {\em The Markup}, 2021.

\bibitem{kodiyan2019overview}
A.~A. Kodiyan, ``An overview of ethical issues in using ai systems in hiring with a case study of amazon’s ai based hiring tool,'' {\em Researchgate Preprint}, pp.~1--19, 2019.

\bibitem{angwin2016machine}
J.~Angwin, J.~Larson, S.~Mattu, and L.~Kirchner, ``Machine bias: There’s software used across the country to predict future criminals. and it’s biased against blacks.,'' {\em ProPublica}, May 2016.

\bibitem{chakraborty2021bias}
J.~Chakraborty, S.~Majumder, and T.~Menzies, ``Bias in machine learning software: Why? how? what to do?,'' in {\em Proceedings of the 29th ACM joint meeting on European software engineering conference and symposium on the foundations of software engineering}, pp.~429--440, 2021.

\bibitem{chen2023fairness}
Z.~Chen, J.~M. Zhang, M.~Hort, M.~Harman, and F.~Sarro, ``Fairness testing: A comprehensive survey and analysis of trends,'' {\em ACM Transactions on Software Engineering and Methodology}, 2023.

\bibitem{chen2022maat}
Z.~Chen, J.~M. Zhang, F.~Sarro, and M.~Harman, ``Maat: a novel ensemble approach to addressing fairness and performance bugs for machine learning software,'' in {\em Proceedings of the 30th ACM joint european software engineering conference and symposium on the foundations of software engineering}, pp.~1122--1134, 2022.

\bibitem{gomez2021winner}
E.~G{\'o}mez, C.~Shui~Zhang, L.~Boratto, M.~Salam{\'o}, and M.~Marras, ``The winner takes it all: geographic imbalance and provider (un) fairness in educational recommender systems,'' in {\em Proceedings of the 44th International ACM SIGIR Conference on Research and Development in Information Retrieval}, pp.~1808--1812, 2021.

\bibitem{oneto2020fairness}
L.~Oneto and S.~Chiappa, ``Fairness in machine learning,'' in {\em Recent trends in learning from data: Tutorials from the inns big data and deep learning conference (innsbddl2019)}, pp.~155--196, Springer, 2020.

\bibitem{zhang2022testsgd}
M.~Zhang, J.~Sun, J.~Wang, and B.~Sun, ``Testsgd: Interpretable testing of neural networks against subtle group discrimination,'' {\em ACM Transactions on Software Engineering and Methodology}, 2022.

\bibitem{buolamwini2018gender}
J.~Buolamwini and T.~Gebru, ``Gender shades: Intersectional accuracy disparities in commercial gender classification,'' in {\em Conference on fairness, accountability and transparency}, pp.~77--91, PMLR, 2018.

\bibitem{chen2022fairness}
Z.~Chen, J.~M. Zhang, M.~Hort, F.~Sarro, and M.~Harman, ``Fairness testing: A comprehensive survey and analysis of trends,'' {\em arXiv preprint arXiv:2207.10223}, 2022.

\bibitem{corbett2018measure}
S.~Corbett-Davies and S.~Goel, ``The measure and mismeasure of fairness: A critical review of fair machine learning,'' {\em arXiv preprint arXiv:1808.00023}, 2018.

\bibitem{cabitza2019wants}
F.~Cabitza and A.~Campagner, ``Who wants accurate models? arguing for a different metrics to take classification models seriously,'' {\em arXiv preprint arXiv:1910.09246}, 2019.

\bibitem{hossin2015review}
M.~Hossin and M.~N. Sulaiman, ``A review on evaluation metrics for data classification evaluations,'' {\em International journal of data mining \& knowledge management process}, vol.~5, no.~2, p.~1, 2015.

\bibitem{galhotra2017fairness}
S.~Galhotra, Y.~Brun, and A.~Meliou, ``Fairness testing: testing software for discrimination,'' in {\em Proceedings of the 2017 11th Joint meeting on foundations of software engineering}, pp.~498--510, 2017.

\bibitem{kearns2018preventing}
M.~Kearns, S.~Neel, A.~Roth, and Z.~S. Wu, ``Preventing fairness gerrymandering: Auditing and learning for subgroup fairness,'' in {\em International conference on machine learning}, pp.~2564--2572, PMLR, 2018.

\bibitem{binns2020apparent}
R.~Binns, ``On the apparent conflict between individual and group fairness,'' in {\em Proceedings of the 2020 conference on fairness, accountability, and transparency}, pp.~514--524, 2020.

\bibitem{wang2022towards}
A.~Wang, V.~V. Ramaswamy, and O.~Russakovsky, ``Towards intersectionality in machine learning: Including more identities, handling underrepresentation, and performing evaluation,'' in {\em Proceedings of the 2022 ACM Conference on Fairness, Accountability, and Transparency}, pp.~336--349, 2022.

\bibitem{guo2021detecting}
W.~Guo and A.~Caliskan, ``Detecting emergent intersectional biases: Contextualized word embeddings contain a distribution of human-like biases,'' in {\em Proceedings of the 2021 AAAI/ACM Conference on AI, Ethics, and Society}, pp.~122--133, 2021.

\bibitem{gohar2023survey}
U.~Gohar and L.~Cheng, ``A survey on intersectional fairness in machine learning: Notions, mitigation, and challenges,'' {\em arXiv preprint arXiv:2305.06969}, 2023.

\bibitem{cabrera2019fairvis}
{\'A}.~A. Cabrera, W.~Epperson, F.~Hohman, M.~Kahng, J.~Morgenstern, and D.~H. Chau, ``Fairvis: Visual analytics for discovering intersectional bias in machine learning,'' in {\em 2019 IEEE Conference on Visual Analytics Science and Technology (VAST)}, pp.~46--56, IEEE, 2019.

\bibitem{tao2022ruler}
G.~Tao, W.~Sun, T.~Han, C.~Fang, and X.~Zhang, ``Ruler: discriminative and iterative adversarial training for deep neural network fairness,'' in {\em Proceedings of the 30th ACM Joint European Software Engineering Conference and Symposium on the Foundations of Software Engineering}, pp.~1173--1184, 2022.

\bibitem{ruf2021towards}
B.~Ruf and M.~Detyniecki, ``Towards the right kind of fairness in ai,'' {\em arXiv preprint arXiv:2102.08453}, 2021.

\bibitem{mehrabi2021survey}
N.~Mehrabi, F.~Morstatter, N.~Saxena, K.~Lerman, and A.~Galstyan, ``A survey on bias and fairness in machine learning,'' {\em ACM computing surveys (CSUR)}, vol.~54, no.~6, pp.~1--35, 2021.

\bibitem{verma2018fairness}
S.~Verma and J.~Rubin, ``Fairness definitions explained,'' in {\em Proceedings of the international workshop on software fairness}, pp.~1--7, 2018.

\bibitem{angwin2016how}
J.~Angwin, J.~Larson, S.~Mattu, and L.~Kirchner, ``How we analyzed the compas recidivism algorithm,'' {\em ProPublica}, May 2016.

\bibitem{dieterich2016compas}
W.~Dieterich, C.~Mendoza, and T.~Brennan, ``Compas risk scales: Demonstrating accuracy equity and predictive parity,'' {\em Northpointe Inc}, vol.~7, no.~4, pp.~1--36, 2016.

\bibitem{flores2016false}
A.~W. Flores, K.~Bechtel, and C.~T. Lowenkamp, ``False positives, false negatives, and false analyses: A rejoinder to machine bias: There's software used across the country to predict future criminals. and it's biased against blacks,'' {\em Fed. Probation}, vol.~80, p.~38, 2016.

\bibitem{saleiro2018aequitas}
P.~Saleiro, B.~Kuester, L.~Hinkson, J.~London, A.~Stevens, A.~Anisfeld, K.~T. Rodolfa, and R.~Ghani, ``Aequitas: A bias and fairness audit toolkit,'' {\em arXiv preprint arXiv:1811.05577}, 2018.

\bibitem{calders2010three}
T.~Calders and S.~Verwer, ``Three naive bayes approaches for discrimination-free classification,'' {\em Data mining and knowledge discovery}, vol.~21, pp.~277--292, 2010.

\bibitem{apyori}
Y.~Mochizuki, ``Apyori.'' \url{https://github.com/ymoch/apyori}, 2019.

\bibitem{guo2023fairrec}
H.~Guo, J.~Li, J.~Wang, X.~Liu, D.~Wang, Z.~Hu, R.~Zhang, and H.~Xue, ``Fairrec: Fairness testing for deep recommender systems,'' {\em arXiv preprint arXiv:2304.07030}, 2023.

\bibitem{misc_diabetes_130-us_hospitals_for_years_1999-2008_296}
J.~Clore, K.~Cios, J.~DeShazo, and B.~Strack, ``{Diabetes 130-US hospitals for years 1999-2008}.'' UCI Machine Learning Repository, 2014.
\newblock {DOI}: https://doi.org/10.24432/C5230J.

\bibitem{dixon2018measuring}
L.~Dixon, J.~Li, J.~Sorensen, N.~Thain, and L.~Vasserman, ``Measuring and mitigating unintended bias in text classification,'' in {\em Proceedings of the 2018 AAAI/ACM Conference on AI, Ethics, and Society}, pp.~67--73, 2018.

\bibitem{zhang2021automatic}
P.~Zhang, J.~Wang, J.~Sun, X.~Wang, G.~Dong, X.~Wang, T.~Dai, and J.~S. Dong, ``Automatic fairness testing of neural classifiers through adversarial sampling,'' {\em IEEE Transactions on Software Engineering}, vol.~48, no.~9, pp.~3593--3612, 2021.

\bibitem{chen2024fairness}
Z.~Chen, J.~Zhang, F.~Sarro, and M.~Harman, ``Fairness improvement with multiple protected attributes: How far are we?,'' in {\em 46th International Conference on Software Engineering (ICSE 2024)}, ACM, 2024.

\end{thebibliography}


\end{document}